\documentclass{article}

\PassOptionsToPackage{numbers, compress}{natbib}

\usepackage[preprint]{neurips_2025}

\usepackage[utf8]{inputenc} 
\usepackage[T1]{fontenc}    
\usepackage{hyperref}       
\usepackage{url}            
\usepackage{booktabs}       
\usepackage{amsfonts}       
\usepackage{nicefrac}       
\usepackage{microtype}      
\usepackage{xcolor}         

\usepackage{xspace}
\usepackage{graphicx}
\usepackage{enumitem}
\usepackage{algorithm}
\usepackage{algorithmic}
\usepackage{amsmath}
\usepackage{subcaption}

\newcommand{\zso}{\textsc{ZeroShotOpt}\xspace}

\title{\textsc{ZeroShotOpt}: Towards Zero-Shot Pretrained\\ Models for Efficient Black-Box Optimization}

\author{
  \textbf{Jamison Meindl}$^{1}$\thanks{Correspondence to: Jamison Meindl <\texttt{jmeindl@mit.edu}>.} \And
  \textbf{Yunsheng Tian}$^{1}$ \And
  \textbf{Tony Cui}$^{1}$ \And 
  \textbf{Veronika Thost}$^{2}$ \And
  \textbf{Zhang-Wei Hong}$^{2}$ \And 
  \textbf{Johannes Dürholt}$^{3}$ \And
  \textbf{Jie Chen}$^{2}$ \And 
  \textbf{Wojciech Matusik}$^{1}$ \And
  \textbf{Mina Konaković Luković}$^{1}$ \\\\
  $^{1}$MIT \quad
  $^{2}$MIT-IBM Watson AI Lab \quad
  $^{3}$Evonik Operations GmbH
}

\begin{document}

\maketitle

\begin{abstract}
    Global optimization of expensive, derivative-free black-box functions requires extreme sample efficiency. While Bayesian optimization (BO) is the current state-of-the-art, its performance hinges on surrogate and acquisition function hyper-parameters that are often hand-tuned and fail to generalize across problem landscapes. We present \zso, a general-purpose, pretrained model for continuous black-box optimization tasks ranging from $2$\,D to $20$\,D. Our approach leverages offline reinforcement learning on large-scale optimization trajectories collected from 12 BO variants. To scale pretraining, we generate millions of synthetic Gaussian process-based functions with diverse landscapes, enabling the model to learn transferable optimization policies. As a result, \zso achieves robust zero-shot generalization on a wide array of unseen benchmarks, matching or surpassing the sample efficiency of leading global optimizers, including BO, while also offering a reusable foundation for future extensions and improvements. Our open-source code, dataset, and model are available at \href{https://github.com/jamisonmeindl/zeroshotopt}{https://github.com/jamisonmeindl/zeroshotopt}.
\end{abstract}

\section{Introduction}

Black-box optimization under tight evaluation budgets is pivotal in many scientific and engineering settings. Since derivatives are often unavailable, classical gradient-based solvers such as Newton’s method or conjugate-gradient are not applicable. Practitioners therefore turn to gradient-free heuristics such as genetic algorithms, simulated annealing, and evolutionary strategies, which often require thousands of evaluations to converge \citep{holland1992adaptation,kirkpatrick1983optimization,hansen2001completely}. When each experiment or simulation is slow or costly, such sample counts become prohibitive. Therefore, we focus on improving continuous black-box optimization under strict evaluation constraints.

Bayesian optimization (BO) \citep{shahriari2015taking} reduces this burden by fitting a probabilistic surrogate (typically a Gaussian process) and selecting queries via a heuristic acquisition function that balances exploration and exploitation.  BO has driven progress in materials discovery \citep{erps2021accelerated}, molecular design \citep{griffiths2020constrained}, clinical prognosis \citep{alaa2018autoprognosis}, and hyper-parameter tuning \citep{turner2021bayesian}. However, BO’s success hinges on hand-chosen kernels, acquisition functions, and hyper-parameters whose optimal settings vary across landscapes and are difficult to tune without expert insight or extra evaluations. 

Consider a researcher maximizing the efficacy of a drug formulation. Each candidate formulation must pass a clinical assay, capping the budget at \emph{fifty} trials. The formulation parameters span dosage, delivery rate, and co-administered compounds, which are continuous parameters with unknown interactions. Standard heuristics exhaust the budget before converging, while BO would succeed only if its kernel and acquisition hyperparameters were tuned. Therefore, there is a need for a low-budget general-purpose global optimizer that can perform well without tuning. 

\begin{figure}[ht]
    \centering
    \includegraphics[width=0.9\columnwidth]{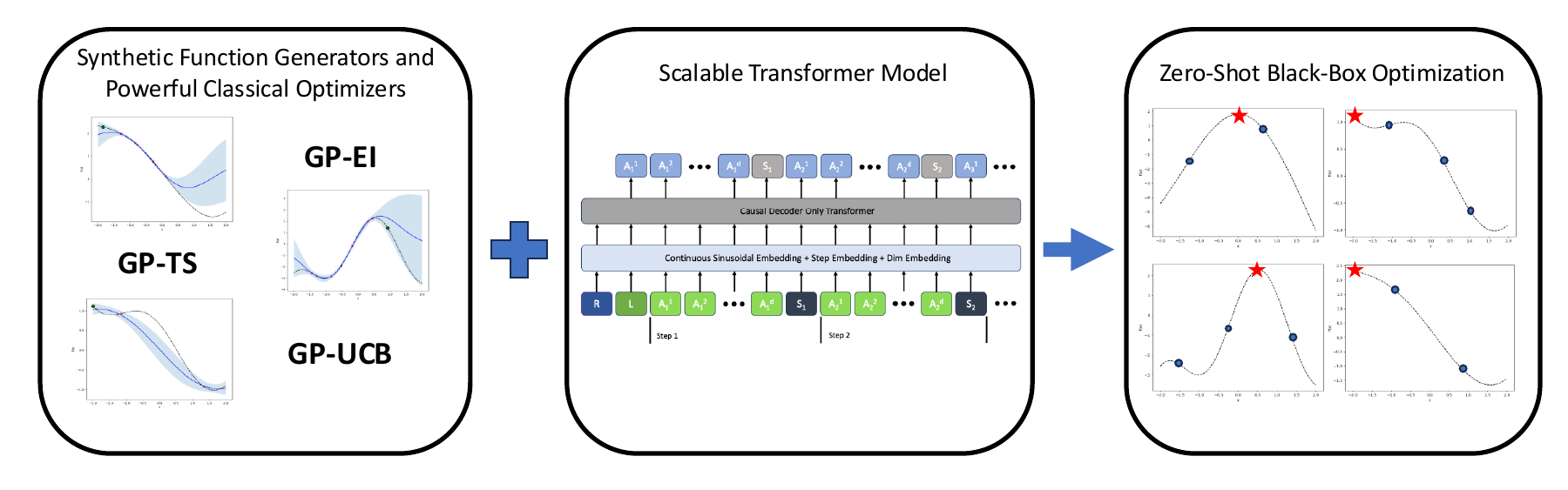}
    \caption{Overall Diagram of \zso. We use a combination of diverse synthetic functions and classical optimizers to train a pretrained transformer model for efficient black-box optimization.}
    \label{fig:intro}
\end{figure}

The large amounts of data and increasing computing power available today have led to the development of pretrained transformer-based models, used in a variety of data-driven decision making scenarios, such as machine translation~\citep{vaswani2017attention}, regression~\citep{song2024omnipred}, and robotic control~\citep{chen2021decision}. Recent approaches have begun exploring using pretrained models for optimization, including both transformer architectures and offline pretraining ~\citep{chen2022towards,krishnamoorthy2022generative,maraval2024end}. However, the full potential of pretrained models in optimization remains untapped, in part due to the limited availability of massive high-quality optimization datasets \citep{song2024position}. Furthermore, demonstrating robust zero-shot generalization on completely unseen problems within low evaluation budgets remains an open challenge, as most pretrained approaches require training on similar data to which the model is tested. Consequently, traditional optimization, such as Bayesian optimization, continues to be the preferred choice in practical applications.

In this paper, we introduce \zso, a \emph{general-purpose pretrained optimizer} for continuous black-box problems up to $20$\,D. We train a $200$\,M parameter transformer-based model using offline reinforcement learning on a diverse dataset of trajectories generated using various BO variants on synthetic GP functions. As a result, \zso demonstrates a robust understanding of optimization dynamics and provides a valuable framework for further optimization improvements. 

In summary, our contributions are as follows.
\begin{enumerate}[leftmargin=*,topsep=0pt,noitemsep]
    \item We develop a \emph{synthetic function generator} based on Gaussian processes and collect \emph{large-scale optimization trajectories} from 12 BO variants as pretraining data. We provide this as an open-source dataset, totaling \( \sim 20\) million synthetic trajectories, providing a valuable large dataset of expert optimization trajectories.
    \item We design and train a single scalable transformer-based model that operates across dimensions from $2$\,D to $20$\,D and up to 50 total evaluations.
    \item Our evaluation demonstrates that \zso achieves \emph{strong zero-shot generalization} performance across a range of unseen global optimization benchmarks, matching the performance of existing state-of-the-art BO methods, while providing the opportunity for further extensions to our method.
\end{enumerate}
We have made our open-source code, complete training and evaluation dataset, and model available at \href{https://github.com/jamisonmeindl/zeroshotopt}{https://github.com/jamisonmeindl/zeroshotopt}.

\section{Related work}

\paragraph{Bayesian optimization (BO)}

BO is a powerful global optimization technique due to its ability to efficiently handle expensive black-box functions by balancing exploration and exploitation with carefully designed acquisition functions, such as Expected Improvement (EI), LogEI~\citep{ament2023unexpected}, Upper Confidence Bound (UCB) and Vizier~\citep{golovin2017google, song2024vizier}. Designed with different principles of trading-off exploration and exploitation, they suit different types of optimization problems. BO has been made accessible by a collection of open-source libraries and software including Spearmint~\citep{snoek2012practical}, BoTorch~\citep{balandat2020botorch}, AutoOED~\citep{tian2021autooed}, SMAC3~\citep{JMLR:v23:21-0888}, HEBO~\citep{Cowen-Rivers2022-HEBO}, Openbox~\citep{jiang2024openbox}, etc. However, it remains a challenge for users to determine the most appropriate configuration for their specific problem setting, as the selection is often heuristic and depends on the problem landscape. 

\paragraph{Offline reinforcement learning}

Decision Transformer \citep{chen2021decision}, which models offline RL as sequence learning, is a primary inspiration for our model architecture. We utilize their idea of conditional reward within transformer-based sequence model as the foundation of our model. Follow-up works on Decision Transformer \citep{yamagata2023q, wu2024elastic} are possible future avenues for improvement. Other insights from offline reinforcement learning \citep{levine2020offlinereinforcementlearningtutorial} could also be used to build upon our methodology. 

\paragraph{Pretrained optimization methods}

Causal transformers have been used for learning to optimize, including BONET~\citep{krishnamoorthy2022generative}, OptFormer~\citep{chen2022towards}, and RIBBO~\citep{song2024reinforced}. These methods train transformer-based models on a variety of synthetic and real-world data and show strong test performance on their test suites. However, these methods primarily rely on specific training sets with similar distributions to their test sets. Further comparisons to \zso are available in Appendix~\ref{app:comp}. The goal of \zso is to provide a novel approach to the challenge of current black-box optimization methods, which suffer from requiring hyperparameter tuning or domain specific adaptation.

Therefore, we propose \zso, a general-purpose, zero-shot, transformer-based optimizer verified on continuous black-box optimization problems ranging from $2$\,D–$20$\,D. Unlike BONET, OptFormer and RIBBO, which each excel only when the test distribution resembles the data they were trained on, \zso is pretrained once on \( \sim 20\) million synthetic trajectories and then can be deployed unchanged on out-of-distribution tests. It retains competitive performance against strong BO baselines on \emph{unseen} synthetic and real-world test suites, while also demonstrating the ability to be fine-tuned to specific domains. Therefore, \zso makes progress towards closing the outstanding gap left by prior transformer-based optimizers by providing a robust out-of-distribution optimization method that provides a base for further improvement.

\section{Approach: \zso}

\textbf{Problem Statement:} For generic black-box global optimization problems, the goal is to solve \(\mathbf{x}^* = \arg\min_{\mathbf{x} \in \mathcal{X}} f(\mathbf{x})\), where \(\mathbf{x} \in \mathbb{R}^d\) is a vector of decision variables, \(\mathcal{X} \subseteq \mathbb{R}^d\) represents the feasible search space, \(f: \mathcal{X} \rightarrow \mathbb{R}\) is the black-box objective function, which is typically expensive to evaluate and lacks gradient information, and \(\mathbf{x}^*\) is the global minimum of \(f(\mathbf{x})\). 

\textbf{Goal:} We aim to develop a pretrained model that serves as a “plug-and-play” optimizer, capable of outperforming traditional black-box optimizers without hyperparameter tuning.

\textbf{Challenge:} Typically, training a model of this nature requires a large dataset of examples as most approaches are based on supervised learning \citep{brown2020language, bordes2024introduction, kim2024openvla}. Therefore, we need expert demonstrations illustrating how to choose evaluation points that minimize functions in low-budget scenarios. Unfortunately, no publicly available dataset provides such demonstrations.

\subsection{Learning to optimize via offline RL}
\label{subsec:opt_mdp}

\textbf{Optimization as sequential decision-making:}
We frame black-box optimization as a sequential decision-making problem, where the optimizer acts as an agent interacting with the black-box function \( f \).  The optimizer observes a set of initial samples:  
\( \{ (\mathbf{x}_1, f(\mathbf{x}_1)), \dots, (\mathbf{x}_m, f(\mathbf{x}_m)) \} \),  
where each \( \mathbf{x}_i \in \mathcal{X} \) is randomly sampled from the search space. At each step \( t \), the optimizer observes all previously evaluated points and their function values,  
\(\{ \mathbf{x}_{1}, f(\mathbf{x}_{1}), \dots, \mathbf{x}_{m+t}, f(\mathbf{x}_{m+t})\} \),  
selects the next evaluation point \( \mathbf{x}_{m+t+1} \) based on its policy \( \pi \), receives the function value  \( f(\mathbf{x}_{m+t+1}) \), and repeats this process until the number of steps $t$ hits the user-defined limit. The agent’s objective is to find a point $\mathbf{x}$ that minimizes the objective value \( f(\mathbf{x}) \) over its interactions with the environment.

\textbf{Implementation:} We adapt Decision Transformer (DT) \citep{chen2021decision}, a scalable, transformer-based, offline RL algorithm, to learn an optimizer policy from static datasets. DT is a transformer-based model that takes the history of states, actions, and trajectory quality as input and predicts actions to achieve trajectories of the specified quality. We define trajectory quality based on normalized regret \( R \) relative to the set of optimization methods run on the same function.

Our adapted DT takes as input past query points \(\mathbf{x}_i\) and their function values \(f(\mathbf{x}_{i})\), along with trajectory regret and length, and predicts the next query point required to generate a trajectory with the given regret and length. Like supervised learning, DT learns from demonstrations in static datasets but differs by conditioning on trajectory regret and length. This is crucial for allowing the model to distinguish between effective and ineffective optimization trajectories. At inference time, we specify zero as the desired regret $R$ to initiate generation, encouraging the model to act as the best methods do. We also specify parameter $L$ as the evaluation budget for the method.

\subsection{Synthetic data generation}
\label{subsec:data}
\begin{figure*}[ht]
    \centering
    \begin{subfigure}{0.16\textwidth}
        \centering
        \includegraphics[width=\textwidth]{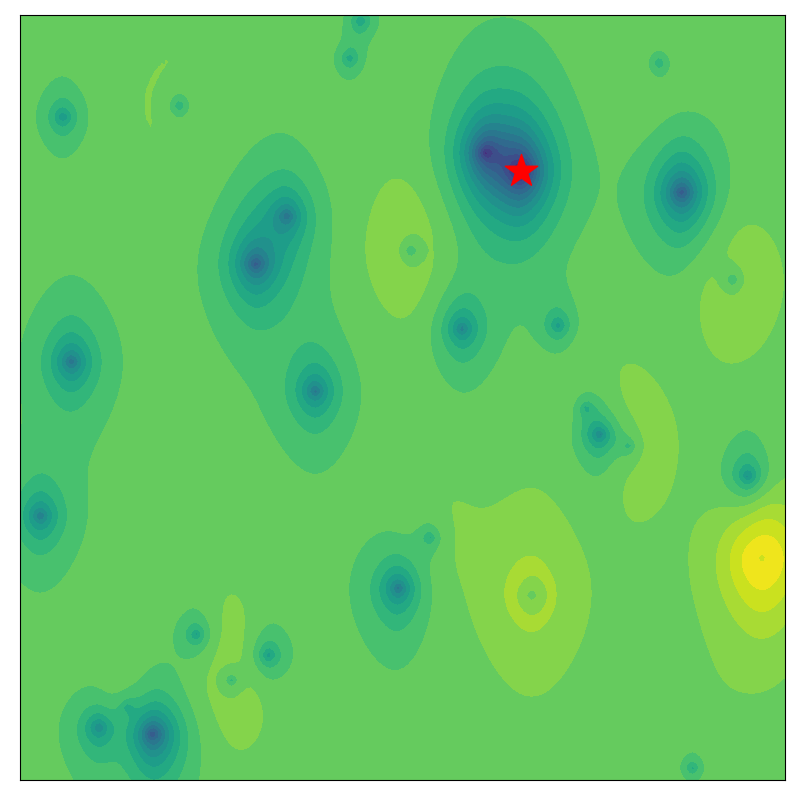}
    \end{subfigure}
    \hfill
    \begin{subfigure}{0.16\textwidth}
        \centering
        \includegraphics[width=\textwidth]{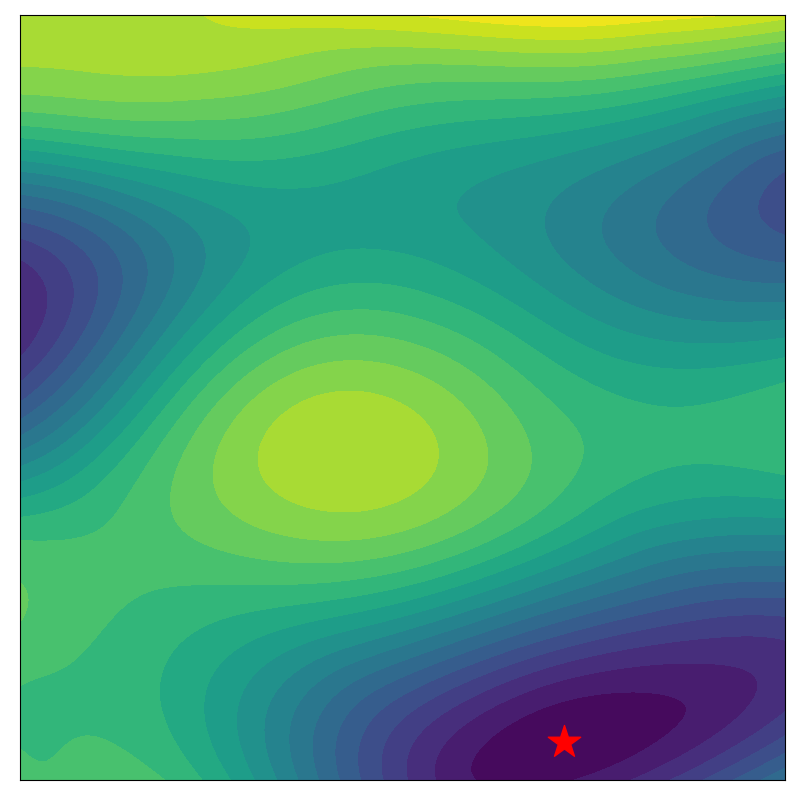}
    \end{subfigure}
    \hfill
    \begin{subfigure}{0.16\textwidth}
        \centering
        \includegraphics[width=\textwidth]{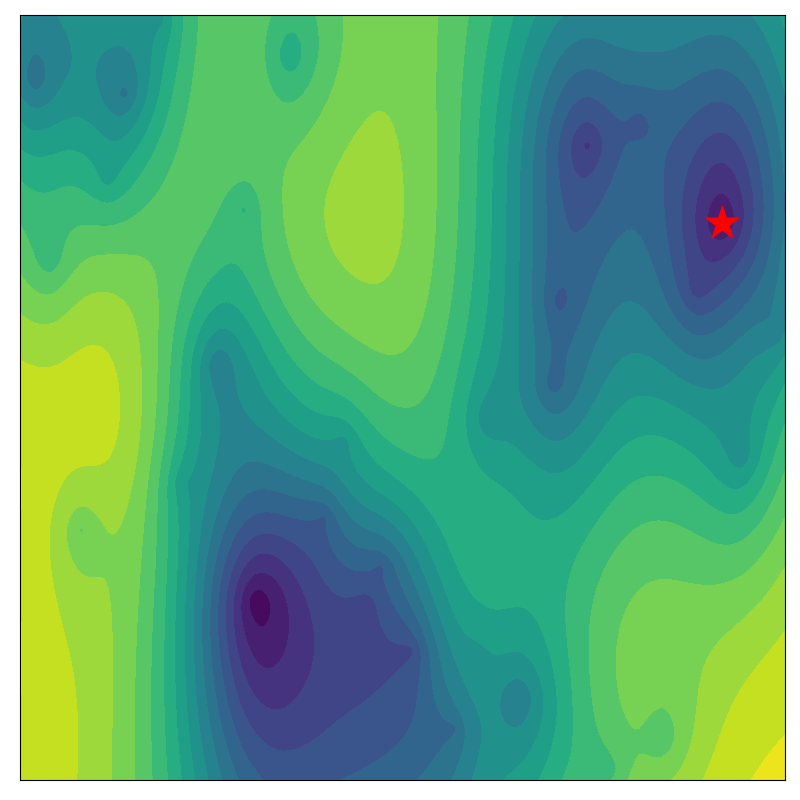}
    \end{subfigure}
    \hfill
    \begin{subfigure}{0.16\textwidth}
        \centering
        \includegraphics[width=\textwidth]{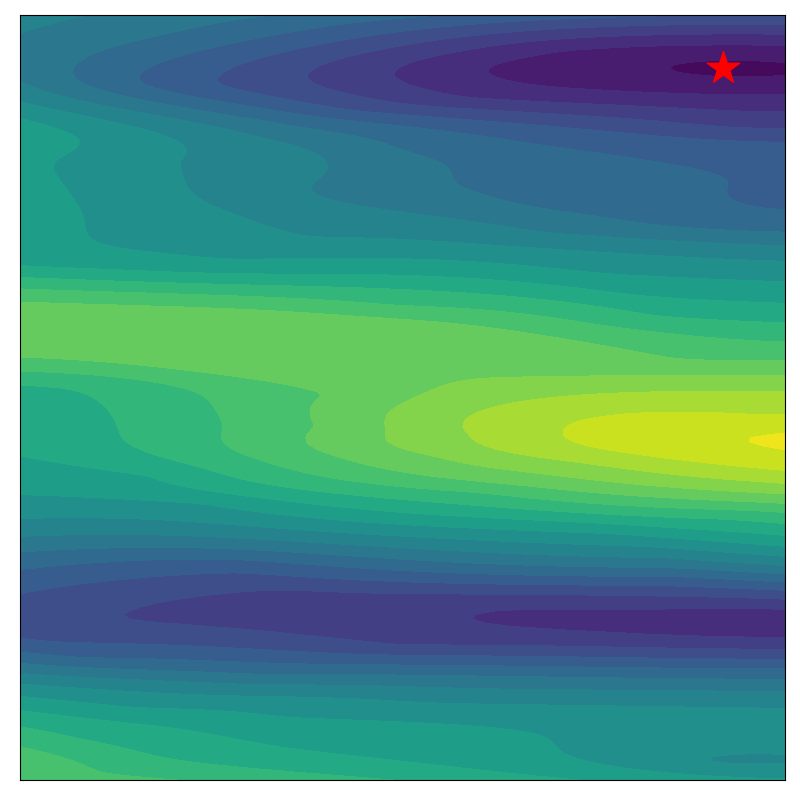}
    \end{subfigure}
    \hfill
    \begin{subfigure}{0.16\textwidth}
        \centering
        \includegraphics[width=\textwidth]{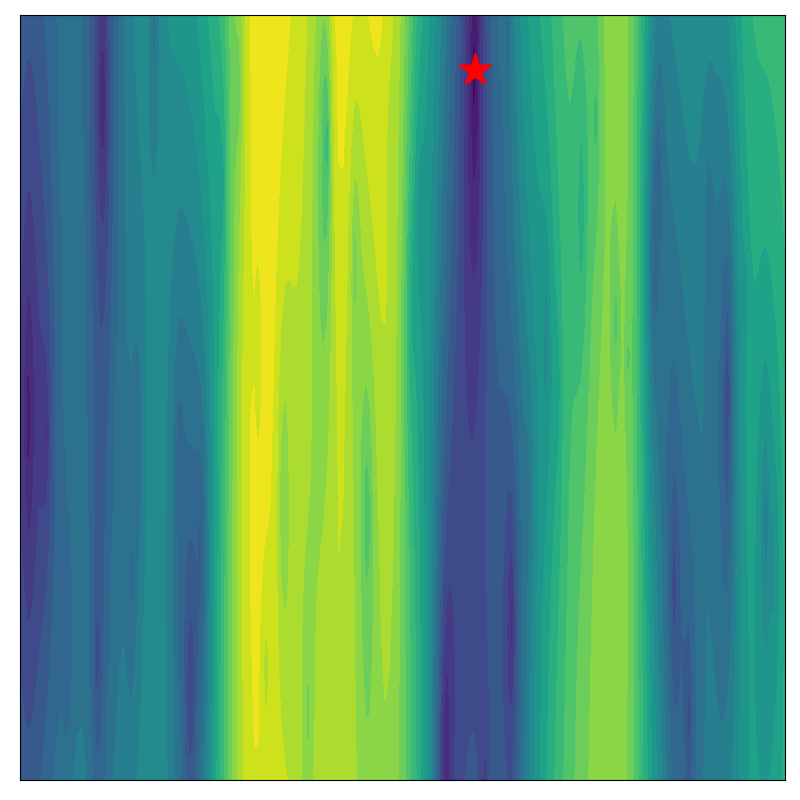}
    \end{subfigure}
    \hfill
    \begin{subfigure}{0.16\textwidth}
        \centering
        \includegraphics[width=\textwidth]{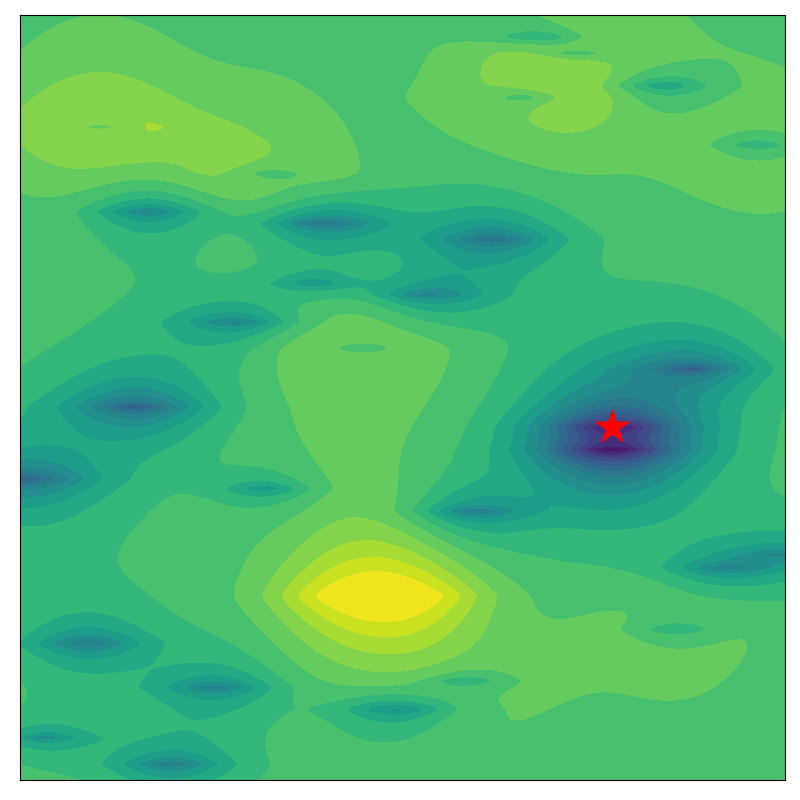}
    \end{subfigure}
    \caption{2D synthetic functions generated by our GP-based function generator with various kernels and parameters. The red star is the global minimum and darker color signifies lower function value.}
    \label{fig:syn_funcs}
\end{figure*}

\textbf{Gaussian process as function generators:} Inspired by \citet{chen2017learning}, we use Gaussian Processes (GPs) as flexible function generators. The intuition here is that GP-based BO methods perform well over a wide variety of function spaces. By basing our training data on GP functions as well, we hope to learn a policy that also adapts well to other function spaces. While GPs may not perfectly represent real-world function spaces, our empirical results show they provide a good basis for learning a policy that works on a wide range of function spaces. To introduce variability, we randomize the kernel type, length scale, and other initialization parameters. Further work in generating synthetic functions could improve the function diversity and therefore results. Figure~\ref{fig:syn_funcs} illustrates example generated functions.

\textbf{Trajectory generation:} For each of the  \(1{,}600{,}000\) synthetic functions generated ranging from $2$\,D to $20$\,D, we run BO with 12 kernel–acquisition variants. These include the Expected Improvement (EI), LogEI \citep{ament2023unexpected}, Upper Confidence Bound (UCB), Joint Entropy Search (JES), Max-value Entropy Search (MES), and Thompson Sampling (TS) acquisition functions with RBF and Matern kernels. Each model is initialized with 10 random points before iteratively fitting the GP and selecting new points using the acquisition function. We generate 12 trajectories per environment, each consisting of 10 initial samples followed by 40 optimization steps, for a total of 50 evaluations.

\subsection{Model architecture}

\begin{figure*}[!h]
    \centering
    \includegraphics[width=0.8\textwidth]{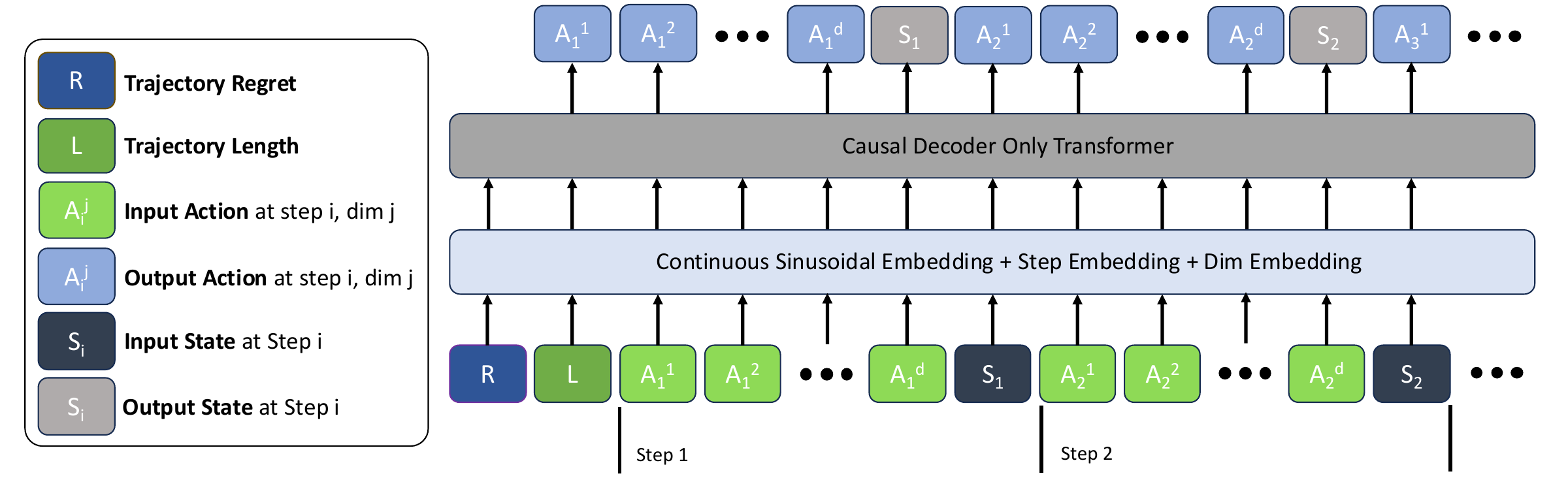}
    \caption{Illustration of the \zso architecture. The model embeds continuous inputs using a fixed sinusoidal embedding and 2 learned positional embeddings. The main model is a causal decoder only transformer, trained on loss on the binned action and state space.}
    \label{fig:pipeline}
\end{figure*}

Our model takes 4 different types of token inputs: regret, length, action (proposed points), and state (function values). The actions are split into individual tokens for each dimension. These are provided as normalized continuous values to a fixed sinusoidal embedding. We add 2 learned positional embeddings, an embedding of the dimension and an embedding of the step. These embeddings are passed on to a causal decoder-only transformer model. This is trained using cross-entropy loss on the binned action and state values, where the normalized inputs are split into 2,000 linearly spaced bins. The main architecture of \zso is illustrated in Figure~\ref{fig:pipeline}.

\subsection{Training and inference}

We train a single model with 200 million parameters on synthetic GP data ranging from $2$\,D to $20$\,D, with step counts of 10, 20, 30, or 40 model steps. We train for a total of 250,000 steps with a batch size of 1,024 trajectories. The model is trained with cross-entropy loss on the binned action and state values. We also test the ability to fine-tune our base model to a new domain, specifically the Hyperparameter Optimization Benchmark (HPO-B) dataset \citep{arango2021hpo}. We split the HPO-B dataset into the specified train and test datasets and generate trajectories on 200 environments from each of the 13 HPO-B classes. We use this small dataset to fine-tune the model to the new domain.

At inference time, we begin by sampling 10 random steps and evaluating these points, as done during dataset generation. We feed this data into the model and iteratively generate 4 proposed actions with the model, along with a predicted state distribution for each action. Similar to OptFormer~\citep{chen2022towards}, we use the expected improvement acquisition function to select an action for that iteration. We then evaluate the action, and feed the new information into the model until the step count is exhausted. We normalize the actions via their allowed range and the state via a probable scaled range. More details on the training and inference process is available in Appendix~\ref{app:impl}.

\section{Experiments}

\subsection{Benchmarks and experimental setup}

We utilize our GP function generator to create a test set of diverse GP functions to test our models in-distribution performance. To test out-of-distribution performance, we use the Virtual Library of Simulated Experiments (VLSE)~\citep{simulationlib} and Black-Box Optimization Benchmark (BBOB)~\citep{elhara2019coco}. These both contain synthetic functions that are traditionally used as standard benchmarks for global black-box optimization. We test our method on $2$\,D, $5$\,D, $10$\,D and $20$\,D function spaces. To evaluate our model on real-world scenarios, we use the HPO-B dataset \citep{arango2021hpo}. HPO-B contains optimization benchmarks built on real-world experiments from OpenML~\citep{vanschoren2014openml} that have been cleaned up for benchmarking. We show results over the 13 search spaces ranging from $2$\,D to $10$\,D. 

We utilize the same global optimizers used to generate training data as our baselines. This includes the following acquisition functions: EI, LogEI, UCB, JES, MES, and TS, all with both RBF and Matern kernels. These baselines were implemented in BoTorch~\citep{balandat2020botorch}. We also compare to other gradient-free optimizers, including Covariance matrix adaptation evolution strategy (CMA-ES), particle swarm optimization (PSO), and differential evolution (DE). We show the performance of each method on these benchmarks using normalized performance. We measure this with $P = (f^* - \min(\{f(\mathbf{x}_{1}), \dots, f(\mathbf{x}_{L})\}))/(f_{m}-f^*)$, where $f^*$ is the global minimum and $f_{m}$ is the median of the initial randomly sampled points. We report the mean performance over tested functions. 

\subsection{Model performance}

We show the mean normalized performance after 50 total evaluations on three datasets in Table~\ref{tab:synthetic-hpob}. \zso consistently achieves top-tier performance across the GP test suite and out-of-distribution test suites (BBOB and VLSE). This is important because \zso was trained on the GP data, so the BBOB and VLSE tests are completely zero-shot. We see that \zso is the top performing method for both synthetic test suites. Additionally, we see that the fine-tuned \zso model performs second best on the HPO-B dataset. We also see the challenge of selecting a BO method here, as the top performing methods are different than on the synthetic datasets. While the original \zso remains competitive with the top BO methods, fine-tuning further turns \zso into a top-performing method.

\begin{table}[ht]
\centering
\caption{Mean normalized performance after 50 total evaluations (10 initial + 40 model steps), averaged over $2$\,D, $5$\,D, $10$\,D and $20$\,D synthetic functions (GP, BBOB \& VLSE) and HPO-B functions ranging from 2D to 10D. We evaluate mean and standard deviation across 5 independent seed splits. }
\resizebox{\textwidth}{!}{
\begin{tabular}{lcccccccc}
\toprule
\multicolumn{1}{c}{} 
  & \multicolumn{2}{c}{\textbf{GP}} 
  & \multicolumn{2}{c}{\textbf{BBOB \& VLSE}}
  & \multicolumn{2}{c}{\textbf{HPO-B}} \\
\cmidrule(r){2-3} \cmidrule(r){4-5} \cmidrule(r){6-7}
\textbf{Method} 
  & Mean $\pm$ SD & Rank 
  & Mean $\pm$ SD & Rank
  & Mean $\pm$ SD & Rank \\
\midrule
\zso        & \textbf{0.647 $\pm$ 0.011} & \textbf{1} 
                            & \textbf{0.881 $\pm$ 0.003} & \textbf{1} 
                            & 0.885 $\pm$ 0.009 & 8 \\
GP-LogEI (Matern)           & 0.629 $\pm$ 0.008 & 2 
                            & 0.878 $\pm$ 0.006 & 2 
                            & 0.900 $\pm$ 0.007 & 5 \\
GP-MES (Matern)             & 0.622 $\pm$ 0.012 & 3 
                            & 0.868 $\pm$ 0.006 & 6 
                            & 0.887 $\pm$ 0.010 & 7 \\
GP-UCB (Matern)             & 0.610 $\pm$ 0.004 & 4 
                            & 0.866 $\pm$ 0.004 & 7 
                            & 0.829 $\pm$ 0.013 & 16 \\
GP-EI (Matern)              & 0.604 $\pm$ 0.004 & 5 
                            & 0.878 $\pm$ 0.004 & 2 
                            & 0.866 $\pm$ 0.007 & 11 \\
GP-UCB (RBF)                & 0.603 $\pm$ 0.011 & 6 
                            & 0.865 $\pm$ 0.006 & 8 
                            & 0.845 $\pm$ 0.011 & 14 \\
GP-LogEI (RBF)              & 0.598 $\pm$ 0.016 & 7 
                            & 0.870 $\pm$ 0.004 & 4 
                            & 0.903 $\pm$ 0.003 & 4 \\
GP-MES (RBF)                & 0.589 $\pm$ 0.007 & 8 
                            & 0.861 $\pm$ 0.005 & 9 
                            & 0.890 $\pm$ 0.008 & 6 \\
GP-EI (RBF)                 & 0.580 $\pm$ 0.008 & 9 
                            & 0.869 $\pm$ 0.002 & 5 
                            & 0.868 $\pm$ 0.004 & 10 \\
GP-JES (Matern)             & 0.518 $\pm$ 0.010 & 10
                            & 0.804 $\pm$ 0.005 & 11 
                            & \textbf{0.915 $\pm$ 0.003} & \textbf{1} \\
GP-JES (RBF)                & 0.512 $\pm$ 0.016 & 11
                            & 0.798 $\pm$ 0.005 & 12 
                            & 0.913 $\pm$ 0.004 & 3 \\
GP-TS (Matern)              & 0.508 $\pm$ 0.013 & 12
                            & 0.795 $\pm$ 0.004 & 14 
                            & 0.844 $\pm$ 0.004 & 15 \\
GP-TS (RBF)                 & 0.506 $\pm$ 0.016 & 13
                            & 0.797 $\pm$ 0.007 & 13 
                            & 0.848 $\pm$ 0.007 & 13 \\
PSO                         & 0.468 $\pm$ 0.009 & 14
                            & 0.790 $\pm$ 0.006 & 15 
                            & 0.856 $\pm$ 0.006 & 12 \\
CMA-ES                      & 0.442 $\pm$ 0.007 & 15
                            & 0.827 $\pm$ 0.007 & 10 
                            & 0.740 $\pm$ 0.014 & 17 \\
DE                          & 0.420 $\pm$ 0.008 & 16
                            & 0.718 $\pm$ 0.005 & 16 
                            & 0.877 $\pm$ 0.005 & 9 \\
\midrule
\zso Fine-Tune 
                            & -- & -- 
                            & -- & -- 
                            & 0.913 $\pm$ 0.002 & 2 \\
\bottomrule
\end{tabular}
}
\label{tab:synthetic-hpob}
\end{table}

\subsection{Runtime results}

We compare the runtime of the best BO method on the synthetic data, GP-LogEI (Matern), to the runtime of \zso across various dimensions in Table~\ref{tab:runtime}. We report the average runtime per evaluation, measured on an NVIDIA H100 GPU using our GP benchmark suite. The runtime of both \zso and GP-LogEI (Matern) increases with higher dimensions, with the exception of GP-LogEI (Matern) at 20D (due to inherent variability in GP fitting and acquisition optimization, smoother acquisition landscapes, or reduced convergence that occurs in higher dimensions). However, \zso remains substantially faster than GP-LogEI (Matern) across all dimensions. 

\begin{table}[t]
\centering
\caption{Average runtime (seconds) per evaluation over different dimensions.}
\begin{tabular}{lcccc}
\toprule
\textbf{Method} & \textbf{2D} & \textbf{5D} & \textbf{10D} & \textbf{20D} \\
\midrule
\zso & 0.368 & 0.778 & 1.889 & 5.556 \\
GP-LogEI (Matern) & 30.597 & 50.030 & 92.865 & 87.752 \\
\bottomrule
\end{tabular}
\label{tab:runtime}
\end{table}

\section{Conclusion, limitations, and future work}

In this work, we proposed \zso, a novel approach for black-box global optimization using a pretrained transformer model trained with offline RL. We formulated the optimization task as a sequential decision-making problem and utilized GPs to create a wide variety of synthetic functions for training. Learning from high-quality trajectories from multiple expert optimizers, our model shows strong performance on unseen optimization problems without the need for task-specific tuning. This work highlights the potential of data-driven pretrained models with GPU acceleration for advancing global optimization and naturally paves the way for promising future extensions.

Our approach is currently limited to continuous, single-objective optimization under 20D, which restricts its applicability to problems involving combinatorial or mixed-integer decision spaces, as well as multi-objective scenarios. Extending the model to tackle these broader categories could significantly enhance its versatility. Additionally, reducing model size through parameter-efficient techniques or pruning could improve deployment efficiency. Lastly, while \zso is competitive with \emph{all} baseline BO variants and generally matches or beats the best performing BO variants, it is occasionally outperformed by the methods with the best variants for individual evaluation distributions. Further improvement beyond the performance of the top BO methods would increase the usefulness of the model.

However, there are many advantages of our transformer-based approach in comparison to BO and opportunities for future work. First, if training data is available, we show that \zso can be fine-tuned to specific domains. Additionally, \zso could be augmented to include semantic information that is difficult to incorporate with BO-based methods. Including information such as historical data from previous similar tests or parameter names or definitions could provide valuable information that models based on \zso could take advantage of. Lastly, scaling up with additional training data from a broader range of functions, as well as utilizing larger networks, is expected to improve performance. Therefore, \zso provides a base that can be utilized to extend to different contexts or quantities of information in a way that BO cannot. Making these extensions possible is a valuable part this work.

\section*{Acknowledgments}
We would like to thank Ashim Sitoula for early experiments on RL and environment generation. This work is supported by Jane Street and the MIT-IBM Watson AI Lab.

\medskip

{
\small
\bibliographystyle{unsrtnat}
\bibliography{reference}

}

\appendix

\newpage
\section{Implementation details}
\label{app:impl}

\subsection{Data generation}

The ability to generate diverse functions, while scaling up in terms of complexity and dimensionality, is crucial to our model's success. Inspired by \citet{chen2017learning}, we use the expressive power of Gaussian Processes (GPs) as a general function generator. Specifically, we randomize the kernel type (\(k(\mathbf{x}, \mathbf{x}')\)) and its length scale to create variability. We randomly sample \(n\) input points \(\{\mathbf{x}_i\}_{i=1}^n \subset \mathcal{X}\) within the search space and assign random objective values \(\{y_i\}_{i=1}^n\). Using these samples, we fit a GP to obtain the posterior mean \(\mu(\mathbf{x})\) and covariance \(k(\mathbf{x}, \mathbf{x}')\), resulting in a posterior function that serves as the training function.

To evaluate the function at a specific point \(\mathbf{x}\), we perform inference on the GP posterior, yielding a sample from the predictive distribution: \(f(\mathbf{x}) \sim \mathcal{N}(\mu(\mathbf{x}), \sigma^2(\mathbf{x}))\), where \(\mu(\mathbf{x})\) is the mean of the posterior and \(\sigma^2(\mathbf{x})\) is the variance derived from the kernel. We sample a single variance input used over the whole environment to generate a continuous function. This approach allows us to efficiently generate training functions with varied and complex behavior. 

In our implementation based on \citet{gpy2014}, we randomly select the kernel type from the set of RBF, Matern 3/2, Matern 5/2, Exponential, Cosine and Quadratic kernels. We also further randomize the kernel by combining up to two kernels together, by adding or multiplying randomly selected kernels. In total, we include 78 different kernel types by combining kernels. We also randomize the length scale between 0.1 and 10, and the number of input points for fitting the GP between $10*d$ and $30*d$, where $d$ is the dimension of the action space. This diverse setup ensures variability and richness in the generated functions. A few examples of a generated functions are shown in Figure~\ref{fig:syn_funcs}.

We use a variety of global optimization algorithms in each environment to generate expert trajectories. We vary the acquisition function and the kernel used to fit each model. We utilize both RBF and matern kernels for each process, with Expected Improvement (EI), Log Expected Improvement (LogEI), Upper Confidence Bound (UCB), Joint Entropy Search (JES), Max-value Entropy Search (MES), and Thompson Sampling (TS) as acquisition functions. We begin by sampling 10 initial points randomly to initialize each model. We then fit the GP model to the set of points and use the specified acquisition function to sample a prospective point. After evaluating that point, we refit the model and continue iteratively. In total, we generate 12 trajectories for each environment. These trajectories contain the 10 initial samples and then a series of points generated by each method. We generate trajectories of length 40 steps, not including the initial samples, for a total of 50 evaluations. We use these trajectories as a baseline for our input regret value and include all as training data for our model. 

We generate trajectories using CPU machines, primarily on an Intel Xeon
Platinum 8260 system. We find that, although BoTorch supports GPU acceleration, the most cost-efficient manner to generate trajectories is to use parallel CPU processes. However, even with optimizations, running our baseline global optimizers is slow, especially for higher dimensions. Therefore, we total $\sim 150,000$ vCPU hours for our data generation. Additionally, we were limited by the amount of trajectories we could generate and more data may help improve the model.

\subsection{Model architecture and training}

\textbf{Trajectory quality:} Regret is defined as the trajectory quality in our paper. We process the regret by the following transformation for $m$ initial samples and $L$ overall steps:
\begin{align}
    R(\mathbf{x}_{1}, f(\mathbf{x}_{1}), \dots, \mathbf{x}_{L}, f(\mathbf{x}_{L})) = \sqrt{\dfrac{\min(\{f(\mathbf{x}_{1}), \dots, f(\mathbf{x}_{L})\}) - f^*_L}{\text{Median}(\{f(\mathbf{x}_{1}), \dots, f(\mathbf{x}_{m})\}) - f^*_L} }
\end{align}
Note that this trajectory quality definition differs from that used in Decision Transformer \citep{chen2021decision}, which employs a return-to-go \(R_t = \sum_{t'=t}^T r_{t'}\) at each step of the trajectory where $r_t$ denotes the reward at times step $t$. In the context of our sequence, unlike in robotic control or Atari games, the state itself reflects the function value, providing direct information about the return. Thus, there is no need for an additional reward signal that differs from the function value. In other words, the state sufficiently conveys the outcomes of prior actions, giving the agent enough context to effectively propose new actions.

\textbf{Training:} Our full training dataset contains $\sim 20,000,000$ total trajectories. We use data augmentation to expand this and enable further generalization. We augment by swapping axes and flipping the action space to provide additional training data. This augmentation greatly expands the trajectory space, particularly for higher dimensions. We also include shortened trajectories (i.e., the first 20 steps of a 40 step trajectory) to expand the dataset. Overall, Table~\ref{tab:counts} contains the number of environments and therefore trajectories of each type before augmentation. We input trajectories of length 10, 20, 30, and 40 to the model, where length does not include the 10 random initial steps.

\begin{table}[ht]
  \centering
  \caption{Number of functions and trajectories for each dimension.}
  \label{tab:counts}
  \begin{tabular}{lrr}
    \toprule
    \textbf{Dimension} & \textbf{Functions} & \textbf{Trajectories} \\
    \midrule
    2D         & 500{,}000     & 6{,}000{,}000 \\
    3D         & 200{,}000     & 2{,}400{,}000 \\
    4D         & 100{,}000     & 1{,}200{,}000 \\
    5D--20D    & 50{,}000 each & 600{,}000 each \\
    \midrule
    \textbf{Total} & \textbf{1{,}600{,}000} & \textbf{19{,}200{,}000} \\
    \bottomrule
  \end{tabular}
\end{table}

Our overall training framework is based on the NanoGPT repository~\citep{Karpathy2022}, which provides a simple and fast GPT implementation, though additional engineering optimizations could further improve efficiency. We adapted this model with our embedding strategy and data pipeline. We train a model with 16 layers, 16 heads, and an embedding dimension of 1024, totaling $\sim200$ million parameters. The action and state space is split into 2,000 evenly spaced bins for our loss calculation. We use trajectories ranging from 2D to 20D and from 10 steps to 40 steps, not including the 10 initial random steps, to train this model. Due to computational constraints and slow generation of higher dimensional data, this dataset contains more data from lower dimensions and fewer steps.

We use the hyperparameters shown in Table~\ref{tab:hyper} to train the model. The base model is trained using the AdamW optimizer using 4 Nvidia H100 GPUs. In total, training takes $\sim 3$ days on this system.

\begin{table}[ht]
  \centering
  \caption{Hyperparameters for model training. The training implementation is largely based on NanoGPT \citep{Karpathy2022}, with adaptations for our specific architecture and data input.}
  \label{tab:hyper}
  \begin{tabular}{ll}
    \toprule
    \textbf{Hyperparameter} & \textbf{Value} \\
    \midrule
    Total Parameters & 200 million \\
    Number of Transformer Heads & 16 \\
    Transformer Embedding Dimension & 1024 \\
    Transformer Layers & 16 \\
    Learning Rate & $6 \times 10^{-4}$ with cosine scaling \\
    Weight Decay & $1 \times 10^{-1}$ \\
    Batch Size & 1024 trajectories \\
    Total Iterations & 250{,}000 \\
    Precision & BF16 \\
    \bottomrule
  \end{tabular}
\end{table}

\subsection{Inference}

\subsubsection{Overall Methodology}
At inference time, we specify zero as the desired regret $R$ to initiate generation, encouraging the agent to act as the best baseline methods do. We then provide the 10 random initial samples, as provided to the global optimization methods to initialize the model. To perform a step, we perform 4 parallel passes of the model at the same time. For each pass, the model generates a distribution across bins for the first dimension. We sample from the distribution using top-p sampling with $p = 0.9$ and convert the bin to a continuous action using the center of the range of the selected bin. We iteratively select actions for each dimension. Once we have sampled the full action dimension for each pass, we use the model to predict a state distribution. Based on these state distributions, we use an expected improvement acquisition function to select the best of the 4 possible actions. We then evaluate the selected action and proceed to the next step. We continue to sample and select actions for the specified number of steps.

\subsubsection{Inference Scaling}
\begin{figure}[ht]
    \centering
    \begin{subfigure}[b]{0.48\textwidth}
        \centering
        \includegraphics[width=\linewidth]{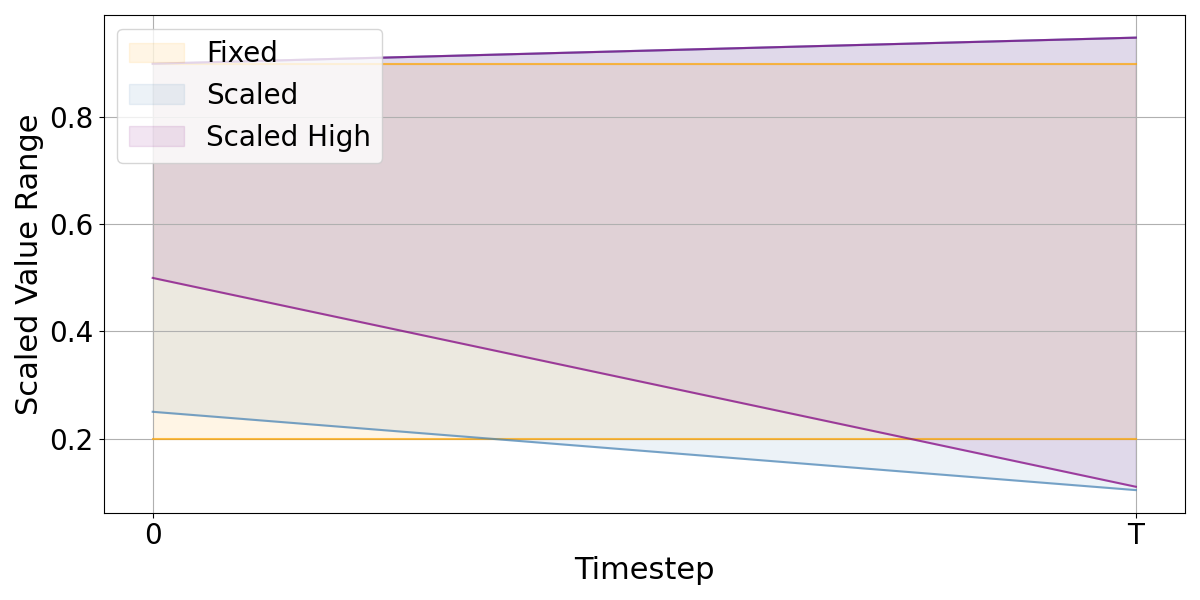}
        \caption{Scaled range of input over steps for each method.}
        \label{fig:inference_a}
    \end{subfigure}
    \hfill
    \begin{subfigure}[b]{0.48\textwidth}
        \centering
        \includegraphics[width=\linewidth]{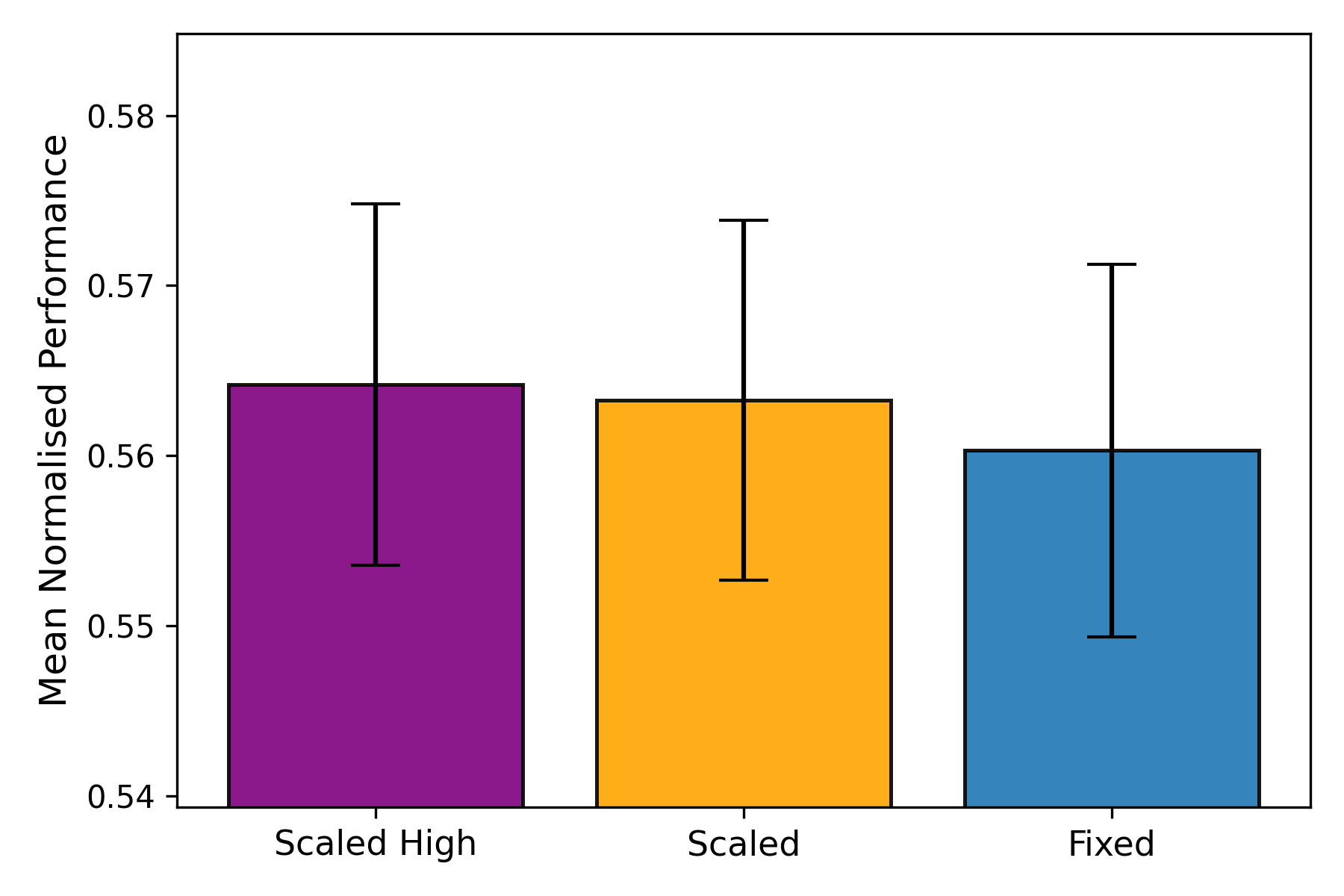}
        \caption{Results on 2D, 5D, 10D, and 20D GP functions.}
        \label{fig:inference_b}
    \end{subfigure}
    \caption{Ablation on scaling strategies during inference.}
    \label{fig:inference_scaling}
\end{figure}

  We show the performance of different inference strategies using the same model in Figure~\ref{fig:inference_scaling}. We scale our states between 0 and 1 during training using the highest and lowest values achieved by any method on the function space, but we do not know these values at inference time. Therefore, we need to develop a scaling strategy for inference. We do this by changing the scaling of the input states at step $t+1$, using parameters $C_u$ and $C_l$ in the following equation:
\[
    S_i' = (S_i - min(S_0, ..., S_{t}))/(max(S_0, ..., S_{t}) - min(S_0, ..., S_{t})) *(1-C_u-C_l) + C_l.
\]

We test various methodologies, from fixed values to scaled values that decrease as the trajectory progresses. We show the ranges of scaled values inputed over T timesteps with 3 methodologies in Figure~\ref{fig:inference_a} and the results of these 3 methodologies in Figure~\ref{fig:inference_b}, which shows a high $C_l$ that decreases with progression performs the best. However, the model is fairly robust to different initializations. 

We define $C_u$ and $C_l$ differently for each method, where $t$ is the current step and $L$ is the number of overall steps. We trial multiple different methodologies for scaling states during inference. These methods vary the way we normalize the state values, given we have less information about the sequence at inference time. 

\textbf{Fixed:} $C_u: 0.1$, $C_l: 0.2$

\textbf{Scaled:} $C_u: 0.05 + 0.05 * \frac{L-t}{L}$, $C_l: 0.1 + 0.15 * \frac{L-t}{L}$

\textbf{Scaled High:} $C_u: 0.05 + 0.05 * \frac{L-t}{L}$, $C_l: 0.1 + 0.4 * \frac{L-t}{L}$

As shown in Figure~\ref{fig:inference_scaling}, we find that \textbf{Scaled High} methodology produces the best results. Therefore, we use this scaling for all presented results. Keeping our scaling configuration consistent across tests allows for flexible model usage and improved utility of our model.

\subsubsection{Inference Runtime}

We show complete results comparing the runtime of \zso to our baseline methods in Table~\ref{tab:full_runtime}. We report the average runtime per evaluation, measured on an NVIDIA H100 GPU using our GP benchmark suite. We see that there is some variance across BO methods due to the nature of different acquisition functions, but \zso is generally faster than most BO methods and can be made more efficient by improved techniques in transformer GPU acceleration. CMA-ES, DE, and PSO are all much faster due to their evolutionary nature, but this comes with much worse performance, which is not preferred for optimizations with low evaluation budgets. 

\begin{table}[t]
\centering
\caption{Average runtime (seconds) per evaluation over different dimensions, sorted by 20D runtime.}
\begin{tabular}{lcccc}
\toprule
\textbf{Method} & \textbf{2D} & \textbf{5D} & \textbf{10D} & \textbf{20D} \\
\midrule
CMA-ES                 & 0.017 & 0.022 & 0.040 & 0.083 \\
PSO                    & 0.032 & 0.043 & 0.090 & 0.208 \\
DE                     & 0.033 & 0.044 & 0.091 & 0.209 \\
GP-JES (RBF)           & 2.012 & 1.972 & 1.888 & 1.905 \\
GP-JES (Matern)        & 2.018 & 2.067 & 2.061 & 2.116 \\
\zso & 0.368 & 0.778 & 1.889 & 5.556 \\
GP-TS (Matern)         & 13.041 & 20.094 & 24.262 & 24.326 \\
GP-TS (RBF)            & 12.460 & 18.486 & 26.779 & 24.905 \\
GP-UCB (Matern)        & 20.088 & 31.973 & 43.356 & 36.811 \\
GP-UCB (RBF)           & 19.016 & 32.738 & 44.413 & 37.279 \\
GP-EI (RBF)            & 15.804 & 37.344 & 57.042 & 46.994 \\
GP-EI (Matern)         & 17.093 & 37.621 & 67.082 & 54.672 \\
GP-LogEI (Matern)      & 30.597 & 50.030 & 92.865 & 87.752 \\
GP-MES (RBF)           & 21.625 & 54.192 & 85.005 & 89.352 \\
GP-LogEI (RBF)         & 29.871 & 54.489 & 85.576 & 94.265 \\
GP-MES (Matern)        & 23.166 & 57.048 & 86.132 & 95.136 \\
\bottomrule
\end{tabular}
\label{tab:full_runtime}
\end{table}

\subsection{Model Scaling}

\begin{figure}
    \centering
    \includegraphics[width=0.75\linewidth]{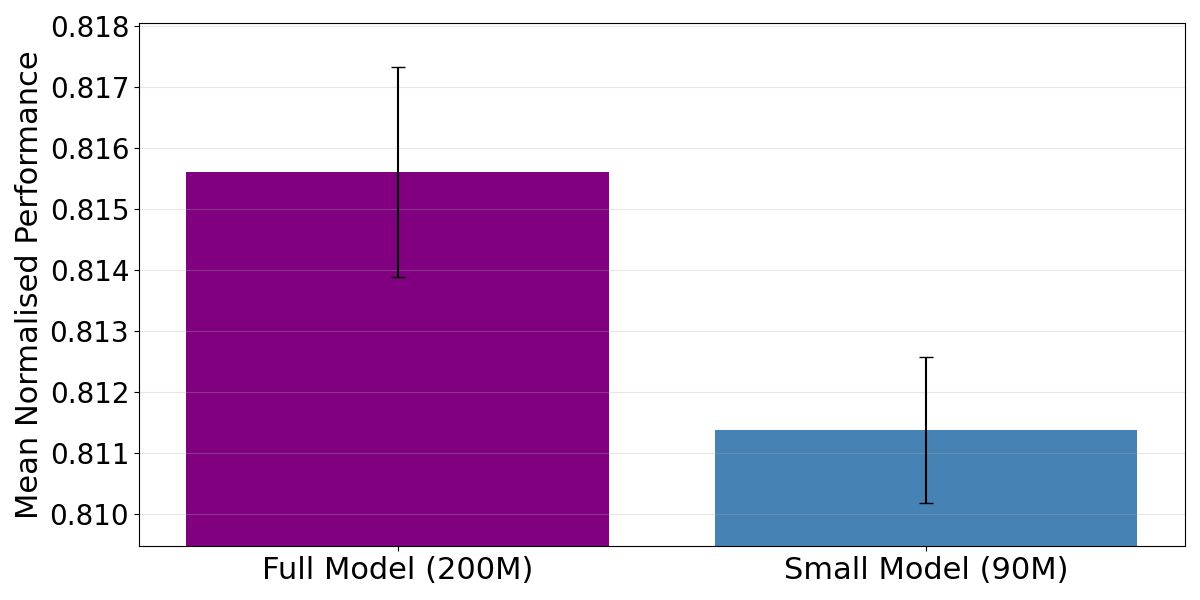}
    \caption{Ablation on Model Size. Mean normalized performance over steps 10, 20, 30 and 40 on $2$\,D, $5$\,D, $10$\,D and $20$\,D BBOB and VLSE functions. We test over 500 functions from each dataset and evaluate standard deviation across 5 independent seed splits.}
    \label{fig:model_size}
\end{figure}

We complete an ablation on model size, suggesting that further increasing the size of \zso could further improve results. We compare the performance of \zso with different model sizes in Figure~\ref{fig:model_size}. In particular, we compare our full-size \zso model with 200M parameters with a smaller model with 90M parameters. This comparison is done without state prediction and the expected improvement acquisition function due to compute requirements. All training is completed with the same number of iterations and training data. We find that the larger model performs better due to improved capacities, suggesting that further scaling may continue to improve results. 

\section{Model Comparison}
\label{app:comp}

We aim provide further comparison to other proposed learning to optimize methods below. The idea of learning to optimize has been studied relatively early in continuous gradient-based optimization \citep{li2016learning,andrychowicz2016learning,chen2022learning}. Many early works reformulate optimization as a sequence prediction problem, training recurrent neural network (RNN) to predict the next point to evaluate. Yet, gradients need to be provided as additional information and they mostly assume in-domain settings \citep{chen2022learning}.
One notable exception are \citet{chen2017learning}, who show that trained RNNs are able to generalize from simple objective functions to a variety of other unseen test functions without gradient information.

\citet{chaybouti2022meta} train a transformer optimizer using online meta RL, but still limited to learning task-specific solvers without generalization to a wide variety of unseen tasks. Similarly, neural acquisition process (NAP)~\citep{maraval2024end} also considers a similar online RL-based formulation of the problem. They propose an additional training objective to explicitly predicting acquisition function values to guide RL exploration more effectively. They also focus on evaluating transfer learning but, different from us, still train individually for different tasks on selected, task-related data.

Pretrained Optimization Model (POM) \citep{li2024pretrained} introduces a population-based model for zero-shot optimization by formulating optimization as an evolutionary algorithm and using meta-learning for training the model. However, they focus on population-based, high-dimensional problems ($>$100 dimensions), which are beyond the scope we are considering.
Given the close connection with reinforcement learning, there are numerous other works that separately study the problem of learning the surrogate \citep{muller2023pfns4bo,wang2024pre} or the acquisition function \citep{volpp2019meta,hsieh2021reinforced}. Large language models have also been applied out of the box for global optimization; for instance, by using in-context learning \citep{lange2024large,liu2024large}.

We also include more detailed comparison between \zso and other pretrained optimizers (BONET, RIBBO, OptFormer, and PFNs4BO) below to further emphasize the differences and impact of \zso.

\textbf{BONET~\citep{krishnamoorthy2022generative}:} BONET applies a causal transformer and RL-inspired pretraining to specific optimization tasks. They introduce a novel approach by synthesizing optimization trajectories by reordering samples in a given dataset, which creates high quality synthetic trajectories. They show promising results on domain-specific extrapolation, outperforming classical methods in their test domain. In comparison to \zso, they focus on domains with higher budgets (in the hundreds) and show results on a small set of high-dimensional problems. Additionally, the study only evaluates domain-specific extrapolation based on a relatively large-scale dataset, instead of tackling unseen optimization problems with just a few dozens of samples. This is an important distinction, as BONET uses more initial points to initialize the model and trains and tests on similar distributions. Therefore, the general problem being solved by BONET has different limitations. Overall, they provide a methodology that advances ideas in transformer-based optimization, but work towards different problems than ones \zso is designed for.

\textbf{OptFormer~\citep{chen2022towards}:} OptFormer is a text-based transformer framework designed for hyperparameter optimization for general problem space beyond continuous. The model is trained on a proprietary hyperparameter optimization database and exploits the textual nature by leveraging hyperparameter names and descriptions. The method shows promise for transformer-based models trained on large datasets by outperforming solutions such as BO on the given test cases, which are drawn from similar distributions as training data. They train a model on a proprietary dataset, BBOB, and HPO-B data. This shows strong results on the test sets of each respective dataset. As their dataset contains a diverse set of functions and their model shows the ability to learn a variety of different algorithms, their contributions show promise towards the idea of a general-purpose transformer-based optimizer. However, the model shows significant performance degradation when tested on distributions different from its training data. Overall, its strong performance on a large dataset marks an important step towards general-purpose transformer-based optimization, but does not truly work as a general purpose optimizer. 

\begin{figure}
    \centering
    \includegraphics[width=0.75\linewidth]{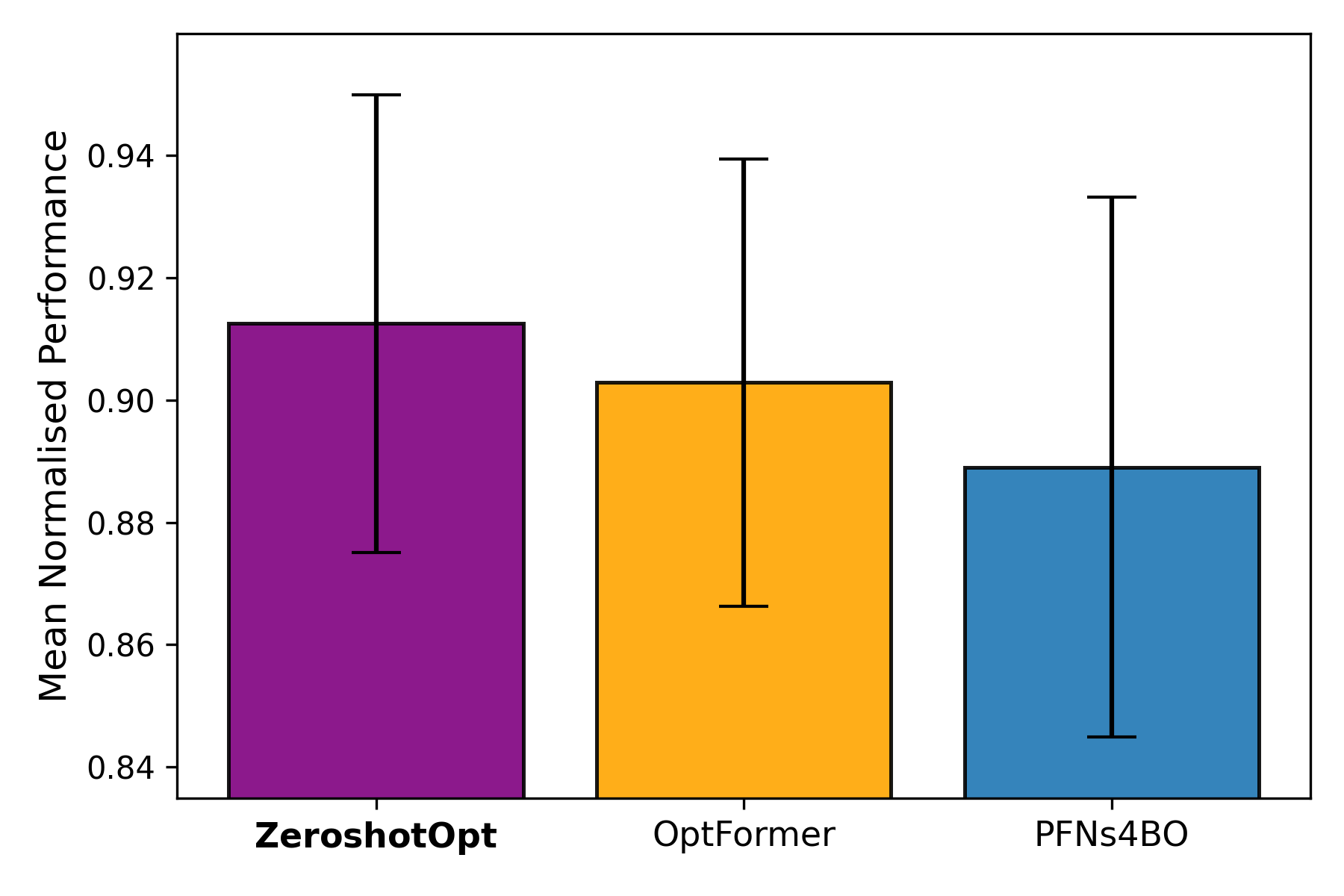}
    \caption{Evaluation Comparison with OptFormer and PFNs4BO. Mean normalized performance over steps 10, 20, 30 and 40 on $2$\,D, $5$\,D, and $10$\,D BBOB functions. We test over 24 functions from each dataset and evaluate standard deviation across 4 independent seed splits.}
    \label{fig:optformer}
\end{figure}

We utilize the pretrained model and code provided by OptFormer to compare to our results on BBOB data. Although we used the default parameters and setup provided in the original documentation, we find the OptFormer model very slow on our hardware, which limited the scope of our experiments. Therefore, we test one environment for each of the 24 BBOB categories for each dimension in $2$\,D, $5$\,D, and $10$\,D. Access to TPU machines may negate these issues. We choose the BBOB tests because although they are contained within the OptFormer training data, we can create general synthetic tests. Therefore, we can limit the impact of including exactly the same data in our test set as OptFormer was trained on. We show the results of our experiments in Figure~\ref{fig:optformer}. We find that \zso outperforms OptFormer, although there is a high standard deviation due to the limited number of tests. Additionally, it is important to note that OptFormer includes BBOB tests within its training data, whereas \zso does not. OptFormer trains on 16 out of the 24 BBOB classes, while specifying 5 examples for out-of-distribution tests. Therefore, this shows the effectiveness of \zso as a general-purpose optimizer that can surpass previous learned optimizers, even when they are trained on task-specific data. 

\textbf{RIBBO~\citep{song2024reinforced}:} RIBBO is a transformer-based framework trained using offline reinforcement learning. While similarly inspired by decision transformer, RIBBO is trained on the benchmark datasets taken from existing optimization tasks. RIBBO shows strong performance and is able to outperform the algorithms it was trained on on many of its test datasets, which is an important contribution. This shows the promising aspect of conditioned transformed-based models. However, RIBBO uses fixed dimensionality, an inflexible tokenization scheme, and a relatively small scale training scheme. Therefore, RIBBO can only evaluate results for small training and evaluation datasets from similar distributions. This differs from \zso, which is designed to be scalable and adapt to different dimension counts and function environments. This scalable nature and use of competitive data generation methods on synthetic functions allows \zso to generalize to different distributions, a feature which is minimal within RIBBO. However, the ideas of conditioned transformer-based models and the improved performance RIBBO shows over baselines on individual datasets provides a step towards outperforming BO in transformer-based optimization.

\textbf{PFNs4BO~\citep{muller2023pfns4bo}:} PFNs4BO use prior-data fitted networks (PFNs) as a surrogate model for Bayesian Optimization. They show the ability to fix to flexible priors, including Gaussian Processes and Bayesian Neural Networks. Similar to us, they train on synthetic datasets with the goal of creating a general-purpose optimizer. They show promising results for their method across a range of test suites, while also allowing the user to add a prior to further improve the model. We show comparison results to \zso and OptFormer in Figure~\ref{fig:optformer}. We see the that transformer-based models outperform PFNs4BO on the BBOB set. However, this method is another promising approach in the learning to optimize space.

\section{Evaluations}
\label{app:eval}

\subsection{Benchmarks}

\textbf{GP:} We use the function generator used for generating training data as the initial baseline for our method. This contains diverse GP functions over a flexible dimension.

\textbf{BBOB~\citep{elhara2019coco}:} The Black-Box Optimization Benchmarking suite was developed as part of the COCO (Comparing Continuous Optimizers) platform to provide a rigorous and standardized environment for evaluating continuous, unconstrained optimization algorithms. BBOB includes 24 benchmark functions that represent a wide range of challenges encountered in real-world optimization, such as separability, multimodality, ill-conditioning, and non-convexity. Each function is parameterized with randomized shifts, scalings, and rotations to prevent algorithms from overfitting to specific patterns. The suite was carefully designed through mathematical constructions and transformations of base functions to create controlled yet diverse test cases. It supports varying dimensions and is widely used in the black-box optimization community. 

\textbf{VLSE~\citep{simulationlib}:} The Virtual Library of Simulation Experiments is a benchmark suite aimed at simulating real-world optimization problems where the objective function is defined by computational simulations rather than closed-form expressions. We use the optimization test problems from this set. 

\textbf{HPO-B~\citep{arango2021hpo}:} The Hyperparameter Optimization Benchmark was created to support fast and reproducible evaluation of hyperparameter optimization methods by providing surrogate models that approximate the behavior of real machine learning training processes. Built from large-scale logging of real hyperparameter tuning runs on algorithms like XGBoost, SVMs, feedforward neural networks, and others across multiple datasets, HPO-B allows researchers to benchmark optimization strategies without incurring the computational cost of retraining models for each evaluation. It enables fair comparisons across different optimizers under controlled experimental conditions. There are a few different evaluation sets provided by HPO-B. We utilize the provided train and test sets for our evaluations.

\subsection{Baselines}

We utilize the same global optimizers used to generate training data as our baselines. This includes the following acquisition functions for BO:

\begin{itemize}
    \item \textbf{Expected Improvement (EI)}: Selects point that is expected to improve upon the current best observation, balancing exploration and exploitation.
    \item \textbf{Log Expected Improvement (LogEI)}: A variant of EI that operates in log space, making it more suitable for objectives with large dynamic ranges or multiplicative noise.
    \item \textbf{Upper Confidence Bound (UCB)}: Prioritizes points with high predicted mean and uncertainty, controlled by a trade-off parameter.
    \item \textbf{Joint Entropy Search (JES)}: Reduces uncertainty about both the location and value of the global minimum by selecting the point expected to maximally reduce the joint entropy of the posterior over the minimum.
    \item \textbf{Max-value Entropy Search (MES)}: Reduces uncertainty about the value of the global minimum by selecting query points that are expected to most reduce the entropy of its posterior distribution.
    \item \textbf{Thompson Sampling (TS)}: Samples functions from the posterior and optimizes them directly, encouraging diverse sampling over time.
\end{itemize}

Each acquisition function is used with both RBF and Matern kernels, providing a good baseline across different smoothness assumptions. We use the default parameterizations of these methods from BoTorch~\citep{balandat2020botorch}, covering a broad range of BO variants.

We also compare to other gradient-free global optimizers:

\begin{itemize}
    \item \textbf{Covariance Matrix Adaptation Evolution Strategy (CMA-ES)}: An evolutionary algorithm that adapts the sampling distribution using covariance information for efficient search.
    \item \textbf{Particle Swarm Optimization (PSO)}: A population-based stochastic optimizer inspired by the social behavior of birds and fish, adjusting candidate solutions based on personal and global bests.
    \item \textbf{Differential Evolution (DE)}: A simple yet powerful method that perturbs candidate solutions using scaled differences between population members.
\end{itemize}

These classical methods typically require thousands of iterations to converge but provide a strong point of comparison to highlight the performance of our model.

\end{document}